\documentclass[11pt]{article}
\usepackage{coling2014}
\usepackage{comment}
\usepackage{times}
\usepackage{url}
\usepackage{latexsym}
\usepackage{amsmath}
\usepackage{amssymb}
\usepackage{multirow}
\usepackage[normalem]{ulem}
\usepackage{color}
\usepackage[usenames,dvipsnames]{xcolor}
\usepackage{enumitem}
\usepackage[lined,boxed,commentsnumbered]{algorithm2e}
\usepackage{graphicx}
\usepackage{epstopdf}
\usepackage{subfig}
\usepackage{mdwlist}
\usepackage{enumitem}
\usepackage{float}

\newtheorem{mydef}{Definition}

\DeclareMathOperator*{\argmax}{arg\,max}

\title{Query-Focused Opinion Summarization for User-Generated Content}

\author{Lu Wang$^{1}$ ~ Hema Raghavan$^{2}$ ~ Claire Cardie$^{1}$ ~ Vittorio Castelli$^{3}$ \\
$^{1}$Department of Computer Science, Cornell University, Ithaca, NY 14853, USA\\
{\tt \{luwang, cardie\}@cs.cornell.edu}\\
$^{2}$LinkedIn, CA, USA\\
{\tt hraghavan@linkedin.com} \\
$^{3}$IBM T. J. Watson Research Center, Yorktown Heights, NY 10598, USA\\
{\tt vittorio@us.ibm.com} \\
}

\begin{document}
\maketitle

\begin{abstract}
\fontsize{10}{11}\selectfont
We present a submodular function-based framework for query-focused opinion summarization. Within our framework, relevance ordering produced by a statistical ranker, and information coverage with respect to topic distribution and diverse viewpoints are both encoded as submodular functions. Dispersion functions are utilized to minimize the redundancy. We are the first to evaluate different metrics of text similarity for submodularity-based summarization methods. By experimenting on community QA and blog summarization, we show that our system outperforms state-of-the-art approaches in both automatic evaluation and human evaluation. A human evaluation task is conducted on Amazon Mechanical Turk with scale, and shows that our systems are able to generate summaries of high overall quality and information diversity.
\end{abstract}

\vspace{-1mm}
\section{Introduction}
\vspace{-1mm}
\blfootnote{\hspace{-0.65cm} This work is licensed under a Creative Commons Attribution 4.0 International Licence. Page numbers and proceedings footer are added by the organisers. Licence details: \url{http://creativecommons.org/licenses/by/4.0/}}

Social media forums, such as social networks, blogs, newsgroups, and community question answering (QA), offer avenues for people to express their opinions as well collect other people's thoughts on topics as diverse as health, politics and software~\cite{Liu:2008:USA:1599081.1599144}.  
However, digesting the large amount of information in long threads on newsgroups, or even knowing
which threads to pay attention to, can be overwhelming. 
A text-based summary that highlights the diversity of opinions on a given topic can lighten this information overload. In this work, we design a submodular function-based framework for opinion summarization on community question answering and blog data.

\vspace{-2mm}
\begin{figure}[ht]
\hspace{-3mm}
    {\fontsize{8.5}{9}\selectfont
    \begin{tabular}{|p{158mm}|}
    \hline
    \textbf{Question}: What is the long term effect of piracy on the music and film industry?\\ \hline \hline
    \textbf{Best Answer}: Rising costs for movies and music. ... If they sell less, they need to raise the price to make up for what they lost. The other thing will be music and movies with less quality. ...\\
	\textbf{Other Answers}:\\
	\textbf{Ans1}: Its bad... really bad. (Just watch this movie and you will find out ... Piracy causes rappers to appear on your computer).\\
	\textbf{Ans2}: By removing the profitability of music \& film companies, piracy takes away their motivation to produce new music \& movies. If they can't protect their copyrights, they can't continue to do business. ...\\
	\textbf{Ans4}: {\color{blue}\textit{It is forcing them to rework their business model, which is a good thing.}} In short, I don't think the music industry in particular will ever enjoy the huge profits of the 90's. ...\\
	\textbf{Ans6}: Please-People in those businesses make millions of dollars as it is!! I don't think piracy hurts them at all!!!
	\\ \hline
	\end{tabular}
	}
	\vspace{-4mm}
    \caption{\fontsize{11}{11}\selectfont 
    Example discussion on Yahoo!~Answers. Besides the best answer, other answers  also contain relevant information (in \textit{italics}). For example, the sentence in {\color{blue} blue} has a contrasting  viewpoint compared to the other answers. }
    \vspace{-4mm}
    \label{fig:yahooqa_example}

\end{figure}

Opinion summarization has previously been applied to restricted domains, such as product reviews~\cite{Hu:2004:MSC:1014052.1014073,Lerman:2009:SSE:1609067.1609124} and news~\cite{Stoyanov:2006:PSC:1610075.1610123}, where the output summary is either presented in a structured way with respect to each aspect of the product or organized along contrastive viewpoints. Unlike those works, we address user generated online data: community QA and blogs. 
These forums use a substantially less formal language than news articles, and at the same time address a much broader spectrum of topics than product reviews. As a result, they present new challenges for automatic summarization. 
For example, Figure~\ref{fig:yahooqa_example} illustrates a sample question from Yahoo!~Answers\footnote{\url{http://answers.yahoo.com/}} along with the answers from different users. The question receives more than one answer, and one of them is selected as the ``best answer" by the asker or other participants. In general, answers from other users also provide relevant information. While community QA successfully pools rich knowledge from the wisdom of the crowd, users might need to seine through numerous posts to extract the information they need.  Hence,  it would be beneficial to summarize answers automatically and present the summaries to users who ask similar questions in the future. In this work, we aim to return a summary that encapsulates different perspectives for a given opinion question and a set of relevant answers or documents.



In our work we assume that there is a central topic (or query) on which a user is seeking diverse opinions. We predict query-relevance through automatically learned statistical rankers. Our ranking function not only aims to find sentences that are on the topic of the query but also ones that are ``opinionated" through the use of several features that indicate subjectivity and sentiment. The relevance score is encoded in a submodular function. Diversity is accounted for by a dispersion function that maximizes the pairwise distance between the pairs of sentences selected. 

Our chief contributions are:\\
(1) We develop a submodular function-based framework for query-focused opinion summarization. To the best of our knowledge, this is the first time that submodular functions have been used to support opinion summarization. We test our framework on two tasks: summarizing opinionated sentences in community QA (Yahoo! Answers) and blogs (TAC-2008 corpus). Human evaluation using Amazon Mechanical Turk shows that our system generates the best summary 57.1\% of the time. On the other hand, the best answer picked by Yahoo! users is chosen only 31.9\% of the time. We also obtain significant higher Pyramid F1 score on the blog task as compared to the system of \newcite{Lin:2011:CSF:2002472.2002537}.
\\
(2) Within our summarization framework, the statistically learned sentence relevance is included as part of our objective function, whereas previous work on submodular summarization~\cite{Lin:2011:CSF:2002472.2002537} only uses ngram overlap for query relevance. Additionally, we use Latent Dirichlet Allocation~\cite{Blei:2003:LDA:944919.944937} to model the topic structure of the sentences, and induce clusterings according to the learned topics. Therefore, our system is capable of generating summaries with broader topic coverage.\\
(3) Furthermore, we are the first to study how different metrics for computing text similarity or dissimilarity affect the quality of submodularity-based summarization methods.
We show empirically that lexical representation-based similarity, such as TFIDF scores, uniformly outperforms semantic similarity computed with WordNet. Moreover, when measuring the summary diversity, topical representation is marginally better than lexical representation, and both of them beats semantic representation.



\vspace{-1mm}
\section{Related Work}
\vspace{-1mm}
Our work falls in the realm of query-focused summarization, where a user asks a question and the system generates a summary of the answers containing pertinent and diverse information. 
A wide range of methods have been investigated, where relevance is often estimated through TF-IDF similarity ~\cite{Carbonell:1998:UMD:290941.291025}, topic signature words~\cite{Lin:2000:AAT:990820.990892} or by learning a Bayesian model over queries and documents~\cite{Daume:2006:BQS:1220175.1220214}. 
Most work only implicitly penalizes summary redundancy, e.g. by downweighting the importance of words that are already selected.

Encouraging diversity of a summary has recently been addressed through submodular functions, which have been applied for multi-document summarization in newswire~\cite{Lin:2011:CSF:2002472.2002537,Sipos:2012:LLS:2380816.2380846}, and comments summarization~\cite{dasgupta-kumar-ravi:2013:ACL2013}. However, these works either ignore the query information (when available) or else use simple ngram matching between the query and sentences. In contrast, we propose to optimize an objective function that addresses both relevance and diversity. 



Previous work on generating opinion summaries mainly considers product reviews~\cite{Hu:2004:MSC:1014052.1014073,Lerman:2009:SSE:1609067.1609124}, and formal texts such as news articles~\cite{Stoyanov:2006:PSC:1610075.1610123} or editorials~\cite{Paul:2010:SCV:1870658.1870665}. 
Mostly, there is no query information, and summaries are formulated in a structured way based on product features or contrastive standpoints. 
Our work is more related to opinion summarization on user-generated content, such as community QA. 
\newcite{Liu:2008:USA:1599081.1599144} manually construct taxonomies for questions in community QA. Summaries are generated by clustering sentences according to their polarity based on a small dictionary. \newcite{Tomasoni:2010:MMA:1858681.1858759} introduce coverage and quality constraints on the sentences, and utilize an integer linear programming framework to select sentences. 

\vspace{-1mm}
\section{Submodular Opinion Summarization}
\vspace{-1mm}
\label{sec:submodular}
In this section, we describe how query-focused opinion summarization can be addressed by submodular functions combined with dispersion functions. We first define our problem. Then we introduce the components of our objective function (Sections~\ref{subsec:relevance}--\ref{subsec:dispersion}). The full objective function is presented in Section~\ref{subsec:objective}. Lastly, we describe a greedy algorithm with constant factor approximation to the optimal solution for generating summaries~(Section~\ref{subsec:greedy}).


A set of documents or answers to be summarized are first split into a set of individual sentences $V=\{s_{1}, \cdots, s_{n}\}$. Our problem is to select a subset $S\subseteq V$ that maximizes a given objective function $f: 2^{V}\rightarrow \mathbb{R}$ within a length constraint:
$S^{\ast}=\underset{S\subseteq V}\argmax  \;f(S)$, subject to $\mid S \mid \leq c$. 
$\mid S \mid$ is the length of the summary $S$, and $c$ is the length limit.

\begin{mydef}
A function $f: 2^{V}\rightarrow \mathbb{R}$ is submodular iff for all $s\in V$ and every $S\subseteq S^{\prime} \subseteq V$, it satisfies $f(S\cup\{s\}) - f(S)\geq f(S^{\prime} \cup\{s\}) - f(S^{\prime})$.
\end{mydef}

Previous submodularity-based summarization work assumes this diminishing return property makes submodular functions a natural fit for summarization and achieves state-of-the-art results on various datasets. 
In this paper, we follow the same assumption and work with non-decreasing submodular functions. 
Nevertheless, they have limitations, one of which is that functions well suited to modeling diversity are not submodular. Recently, \newcite{dasgupta-kumar-ravi:2013:ACL2013} proved that diversity can nonetheless be encoded in well-designed \textit{dispersion functions} which still maintain a constant factor approximation when solved by a greedy algorithm.




Based on these considerations, we propose an objective function $f(S)$ mainly considering three aspects: \textit{relevance} (Section~\ref{subsec:relevance}), \textit{coverage} (Section~\ref{subsec:coverage}), and \textit{non-redundancy} (Section~\ref{subsec:dispersion}). Relevance and coverage are encoded in a non-decreasing submodular function, and non-redundancy is enforced by maximizing the dispersion function.

\vspace{-1mm}
\subsection{Relevance Function}
\label{subsec:relevance}
\vspace{-1mm}
We first utilize statistical rankers to produce a preference ordering of the candidate answers or sentences. 
We choose ListNet~\cite{Cao:2007:LRP:1273496.1273513}, which has been shown to be effective in many information retrieval tasks, as our ranker. We use the implementation from Ranklib~\cite{citeulike:9435511}.

Features used in the ranking algorithm are summarized in Table~\ref{tab:feature_ranking}. All features are normalized by standardization. Due to the length limit, we cannot provide the full results on feature evaluation. Nevertheless, we find that ranking candidates by TFIDF similarity or key phrases overlapping with the query can produce comparable results with using the full feature set (see Section~\ref{sec:results}).

We take the ranks output by the ranker, and define the relevance of the current summary $S$ as: {\small $r(S)=\sum_{i}^{|S|}\sqrt{rank_{i}^{-1}}$}, where $rank_{i}$ is the rank of sentence $s_{i}$ in $V$. For QA answer ranking, sentences from the same answer have the same ranking. The function $r(S)$ is our first submodular function.

\begin{table}
\centering
    {\fontsize{8.5}{8.5}\selectfont
    \setlength{\baselineskip}{0pt}
    \begin{tabular}{|l|l|}
    \hline
    {\bf Basic Features}											&     {\bf Sentiment Features}\\ \hline
    - answer position in all answers/sentence position in blog	&	- number/portion of sentiment words from a lexicon (Section~\ref{subsec:coverage}) \\
    - length of the answer/sentence								& 	- if contains sentiment words with the same polarity as\\
    - length is less than 5 words								&	sentiment words in query \\
	
	\hline								
    {\bf Query-Sentence Overlap Features} 						& {\bf Query-Independent Features} \\ \hline
    - unigram/bigram TF/TFIDF similarity with query	& 	- unigram/bigram TFIDF similarity with cluster centroid\\
    - number of key phrases in the query that appear in the& 	- sumBasic score~\cite{nenkova-vanderwende-2005}\\
    	sentence. A model similar to that described in 	&	- number of topic signature words~\cite{Lin:2000:AAT:990820.990892}\\
    	~\cite{DBLP:conf/naacl/LuoRCMF13} was applied to detect key phrases. 	&    - JS divergence with cluster\\    
    \hline
    
    \end{tabular}
    }
    \vspace{-4mm}
    \caption{\fontsize{11}{11}\selectfont Features used for candidate ranking. We use them for ranking answers in both community QA and blogs.}
    \vspace{-5mm}
    \label{tab:feature_ranking}
\end{table}

\vspace{-1mm}
\subsection{Coverage Functions}
\label{subsec:coverage}
\vspace{-1mm}

\noindent
\textbf{Topic Coverage.}
This function is designed to capture the idea that a comprehensive opinion summary should provide thoughts on distinct aspects. Topic models such as Latent Dirichlet Allocation (LDA)~\cite{Blei:2003:LDA:944919.944937} and its variants are able to discover hidden topics or aspects of document collections, and thus afford a natural way to cluster texts according to their topics. Recent work~\cite{Xie13} shows the effectiveness of utilizing topic models for newsgroup document clustering. We first learn an LDA model from the data, and treat each topic as a cluster. We estimate a sentence-topic distribution $\vec{\theta}$ for each sentence, and assign the sentence to the cluster $k$ corresponding to the mode of the distribution (i.e.,  $k=\argmax_{i}\theta_{i}$). This naive approach produces comparable clustering performance to the state-of-the-art according to~\cite{Xie13}. $\mathcal{T}$ is defined as the clustering induced by our algorithm on the set $V$. The topic coverage of the current summary $S$ is defined as $t(S)=\sum_{T\in \mathcal{T}}\sqrt{\left| S\cap T \right|} $.  From the concavity of the square root it follows that sets $S$ with uniform coverages of topics are preferred to sets with skewed coverage. 

\noindent
\textbf{Authorship Coverage.}
This term encourages the summarization algorithm to select sentences from different authors. Let $\mathcal{A}$ be the clustering induced by the sentence to author relation. In community QA, sentences from the answers given by the same user belong to the same cluster. Similarly, sentences from blogs with the same author are in the same cluster. The authorship score is defined as $a(S)=\sum_{A\in \mathcal{A}}\sqrt{\left| S\cap A \right|}$.

\noindent
\textbf{Polarity Coverage.}
The polarity score encourages the selection of summaries that cover both positive and negative opinions. We categorize each sentence simply by counting the number of polarized words given by our lexicon. A sentence belongs to a positive cluster if it has more positive words than negative ones, and vice versa. If any negator co-occurs with a sentiment word (e.g. within a window of size 5), the sentiment is reversed.\footnote{There exists a large amount of work on determining the polarity of a sentence~\cite{Pang:2008:OMS:1454711.1454712} which can be employed for polarity clustering in this work. We decide to focus on summarization, and estimate sentence polarity through sentiment word summation~\cite{Yu+Hatzivassiloglou:03a}, though we do not distinguish different sentiment words.} The polarity clustering $\mathcal{P}$ thus have two clusters corresponding to positive and negative opinions. The score is defined as $p(S)=\sum_{P\in \mathcal{P}}\sqrt{\mid S\cap P \mid}$. 
Our lexicon consists of MPQA lexicon~\cite{Wilson:2005:RCP:1220575.1220619}, General Inquirer~\cite{stone66}, and SentiWordNet~\cite{Esuli2006sentiwordnet}. 
Words with conflicting sentiments from different lexicons are removed.

\noindent
\textbf{Content Coverage.}
Similarly to~\newcite{Lin:2011:CSF:2002472.2002537} and~\newcite{dasgupta-kumar-ravi:2013:ACL2013}, we use the following function to measure content coverage of the current summary $S$: $c(S)=\sum_{v\in V} \min (cov(v,S), \theta \cdot cov(v,V))$, where $cov(v,S)=\sum_{u\in S} sim(v, u)$. 
We experiment with two types of similarity functions. One is a Cosine TFIDF similarity score. The other is a WordNet-based semantic similarity score between pairwise dependency relations from two sentences~\cite{dasgupta-kumar-ravi:2013:ACL2013}. Specifically, $sim_{Sem}(v, u)=\sum_{rel_{i}\in v, rel_{j}\in u} WN(a_{i}, a_{j})\times WN(b_{i}, b_{j})$, where $rel_{i}=(a_{i}, b_{i}), rel_{j}=(a_{j}, b_{j})$, $WN(w_{i}, w_{j})$ is the shortest path length. 
All scores are scaled onto $[0,1]$.


\vspace{-1mm}
\subsection{Dispersion Function}
\label{subsec:dispersion}
\vspace{-1mm}
Summaries should contain as little redundant information as possible. We achieve this by adding an additional term to the objective function, encoded by a dispersion function. 
Given a set of sentences $S$, a complete graph is constructed with each sentence in $S$ as a node. The weight of each edge ($u, v$) is their dissimilarity $d^{\prime}(u, v)$. Then the distance between any pair of $u$ and $v$, $d(u, v)$, is defined as the total weight of the shortest path connecting $u$ and $v$.\footnote{This definition of distance is used to produce theoretical guarantees for the greedy algorithm described in Section~\ref{subsec:greedy}.} We experiment with two forms of dispersion function~\cite{dasgupta-kumar-ravi:2013:ACL2013}: (1) $h_{sum}=\sum_{u,v\in V, u\neq v} d (u,v)$, and (2) $h_{min}=\min_{u,v\in V, u\neq v} d (u,v)$.

%
%
%
%

Then we need to define the dissimilarity function $d^{\prime}(\cdot, \cdot)$. There are different ways to measure the dissimilarity between sentences~\cite{Mihalcea:2006:CKM:1597538.1597662,agirre-EtAl:2012:STARSEM-SEMEVAL}. In this work, we experiment with three types of dissimilarity functions. 

\noindent
\textbf{Lexical Dissimilarity.}
This function is based on the well-known Cosine similarity score using TFIDF weights. Let $sim_{tfidf} (u,v)$ be the Cosine similarity between $u$ and $v$, then we have $d^{\prime}_{Lex}(u,v)=1-sim_{tfidf} (u,v)$.

\noindent
\textbf{Semantic Dissimilarity.}
This function is based on the semantic meaning embedded in the dependency relations. $d^{\prime}_{Sem}(u,v)=1-sim_{Sem}(v, u)$, where $sim_{Sem}(v, u)$ is the semantic similarity used in content coverage measurement in Section~\ref{subsec:coverage}.

\noindent
\textbf{Topical Dissimilarity.}
We propose a novel dissimilarity measure based on topic models. \newcite{Celikyilmaz:2010:LBS:1867767.1867768} show that estimating the similarity between query and passages by using topic structures can help improve the retrieval performance. As discussed in the topic coverage in Section~\ref{subsec:coverage}, each sentence is represented by its sentence-topic distributions estimated by LDA. For candidate sentence $u$ and $v$, let their topic distributions be $P_{u}$ and $P_{v}$. Then the dissimilarity between $u$ and $v$ can be defined as: {\small $d^{\prime}_{Topic}(u,v)=JSD(P_{u}||P_{v})=\frac{1}{2} (\sum_{i}P_{u}(i)\log_{2} \frac{P_{u}(i)}{P_{a}(i)}+\sum_{i}P_{v}(i)\log_{2} \frac{P_{v}(i)}{P_{a}(i)})$}
where {\small $P_{a}(i)=\frac{1}{2}\left(P_u(i) + P_v(i)\right)$}.

\vspace{-1mm}
\subsection{Full Objective Function}
\label{subsec:objective}
\vspace{-1mm}
The objective function takes the interpolation of the submodular functions and dispersion function:

\vspace{-3mm}
{\small
\begin{equation}
\label{eq:objective}
\mathcal{F}(S)=r(S)+\alpha t(S) +\beta a(S) +\gamma p(S) + \eta c(S) + \delta h(S).
\end{equation}
}
\vspace{-7mm}

The coefficients $\alpha, \beta, \gamma, \eta, \delta$ are non-negative real numbers and can be tuned on a development set.\footnote{The values for the coefficients are $5.0, 1.0, 10.0, 5.0, 10.0$ for $\alpha, \beta, \gamma, \eta, \delta$, respectively, as tuned on the development set.} Notice that each summand except $h(S)$ is a non-decreasing, non-negative, and submodular function, and summation preserves monotonicity, non-negativity, and submodularity. Dispersion function $h(s)$ is either $h_{sum}$ or $h_{min}$ as introduced previously.

%
%
%

\vspace{-1mm}
\subsection{Summary Generation via Greedy Algorithm}
\label{subsec:greedy}
\vspace{-1mm}
Generating the summary that maximizes our objective function in Equation~\ref{eq:objective} is NP-hard~\cite{Chandra:1996:FDR:645898.756652}. We choose to use a greedy algorithm that guarantees to obtain a constant factor approximation to the optimal solution~\cite{citeulike:416655,dasgupta-kumar-ravi:2013:ACL2013}. Concretely, starting with an empty set, for each iteration, we add a new sentence so that the current summary achieves the maximum value of the objective function. 
In addition to the theoretical guarantee, existing work~\cite{McDonald:2007:SGI:1763653.1763720} has empirically shown that classical greedy algorithms usually works near-optimally.

\section{Experimental Setup}
\vspace{-1mm}
\subsection{Opinion Question Identification}
\label{subsec:opinionQuestion}
\vspace{-1mm}
We first build a classifier to automatically detect opinion oriented questions in Community QA; questions in the blog dataset are all opinionated. 
Our opinion question classifier is trained on two opinion question datasets: (1) the first, from \newcite{DBLP:conf/emnlp/LiLA08}, contains 646 opinionated and 332 objective questions; (2) the second dataset, from~\newcite{conf/aaai/AmiriZC13}, consists of 317 implicit opinion questions, such as ``\textit{What can you do to help environment?}", and 317 objective questions. We train a RBF kernel based SVM classifier to identify opinion questions, which achieves F1 scores of $0.79$ and $0.80$ on the two datasets when evaluated using 10-fold cross-validation (the best F1 scores reported are $0.75$ and $0.79$).

\vspace{-1mm}
\subsection{Datasets}
\label{subsec:datasets}
\vspace{-1mm}
\noindent
\textbf{Community QA Summarization: Yahoo! Answers.}
We use the Yahoo! Answers dataset from Yahoo! \textit{Webscope$^{TM}$} program,\footnote{\url{http://sandbox.yahoo.com/}} which contains 3,895,407 questions. We first run the opinion question classifier to identify the opinion questions. For summarization purpose, we require each question having at least 5 answers, with the average length of answers larger than 20 words. This results in 130,609 questions. 

To make a compelling task, we reserve questions with an average length of answers larger than 50 words as our test set for both ranking and summarization; all the other questions are used for training. As a result, we have 92,109 questions in the training set for learning the statistical ranker, and 38,500 in the test set. The category distribution  of training and test questions (Yahoo! Answers organizes the questions into predefined categories) are similar. 10,000 questions from the training set are further reserved as the development set. Each question in the Yahoo! Answers dataset has a user-voted best answer. These best answers are used to train the statistical ranker that predicts relevance. Separate topic models are learned for each category, where the category tag is provided by Yahoo! Answer.

\noindent
\textbf{Blog Summarization: TAC 2008.}
We use the TAC 2008 corpus~\cite{Dang:2008b}, which consists of 25 topics. 23 of them are provided with human labeled nuggets, which TAC used in human evaluation. TAC also provides snippets (i.e., sentences) that are frequently retrieved by participant systems or identified as relevant by human annotators. We do not assume those snippets are known to any of our systems.

\vspace{-1mm}
\subsection{Comparisons}
\vspace{-1mm}
For both opinion summarization tasks, we compare with (1) the approach by~\newcite{dasgupta-kumar-ravi:2013:ACL2013}, and (2) the systems from~\newcite{Lin:2011:CSF:2002472.2002537} with and without query information. The sentence clustering process in \newcite{Lin:2011:CSF:2002472.2002537} is done by using CLUTO~\cite{karypis2003cluto}. For the implementation of systems in \newcite{Lin:2011:CSF:2002472.2002537} and \newcite{dasgupta-kumar-ravi:2013:ACL2013}, we always use the parameters reported to have the best performance in their work.

For cQA summarization, we use the \textbf{best answer} voted by the user as a baseline. Note that this is a strong baseline since all the other systems are unaware of which answer is the best. For blog summarization, we have three additional baselines -- the \textbf{best systems} in TAC 2008~\cite{uiuc,polyu}, top sentences returned by our \textbf{ranker}, a baseline produced by TFIDF similarity and a lexicon (henceforth called \textbf{TFIDF$+$Lexicon}). In TFIDF$+$Lexicon, sentences are ranked by the TFIDF similarity with the query, and then sentences with sentiment words are selected in sequence. This baseline aims to show the performance when we only have access to lexicons without using a learning algorithm.


\section{Results}
\label{sec:results}
\vspace{-1mm}
\subsection{Evaluating the Ranker}
\vspace{-1mm}
We evaluate our ranker (described in Section~\ref{subsec:relevance}) on the task of best answer prediction.
Table~\ref{tab:bestanswer_prediction} compares the average precision and mean reciprocal rank (MRR) of our method to those of three baselines, 
(1) where answers are ranked randomly (\textbf{Baseline (Random)}), (2) by length (\textbf{Baseline (Length)}), and (3) by Jensen Shannon Divergence (\textbf{JSD}) with all answers. 
We expect that the best answer is the one that covers the most information, which is likely to have a smaller JSD. Therefore, we use JSD to rank answers in the ascending order. Table~\ref{tab:bestanswer_prediction} manifests that our ranker outperforms all the other methods.
%

\vspace{-3mm}
\begin{table}[ht]
\centering
    {\fontsize{8.5}{8.5}\selectfont
	\begin{tabular}{|l|c|c|c|c|}
    \hline
    & \textbf{Baseline (Random)} & \textbf{Baseline (Length)} & \textbf{JSD} & \textbf{Ranker (ListNet)} \\ \hline
    Avg Precision & 0.1305 & 0.2834 & 0.4000 & \textbf{0.5336}\\ \hline
    MRR & 0.3403 & 0.4889 & 0.5909 & \textbf{0.6496}\\ \hline
	\end{tabular}
	}
	\vspace{-3mm}
    \caption{\fontsize{11}{11}\selectfont Performance for best answer prediction. Our ranker outperforms the three baselines.}
    \vspace{-7mm}
    \label{tab:bestanswer_prediction}
\end{table}


\subsection{Community QA Summarization}
\label{sec:mturk}

\noindent
\textbf{Automatic Evaluation.} 
Since human written abstracts are not available for the Yahoo! Answers dataset, we adopt the Jensen-Shannon divergence (JSD) to measure the summary quality. 
%
Intuitively, 
a smaller JSD implies that the summary covers more of the content in the answer set. \newcite{Louis:2013:AAM:2483810.2483812} report that JSD has a strong negative correlation (Spearman correlation $=-0.737$) with the overall summary quality for multi-document summarization (MDS) on news articles and blogs. 
Our task is similar to MDS. Meanwhile, the average JSD of the best answers in our test set is smaller than that of the other answers (0.39 vs. 0.49), with an average length of 103 words compared with 67 words for the other answers. 
Also, on the blog task (Section~\ref{subsec:blog}), the top two systems by JSD also have the top two ROUGE scores (a common metric for summarization evaluation when human-constructed summaries are available). 
Thus, we conjecture that JSD is a good metric for community QA summaries.


Table~\ref{tab:JSD_yahoo_main} (left) shows that our system using a content coverage function based on Cosine using TFIDF weights, and a dispersion function ($h_{sum}$) based on lexicon dissimilarity and 100 topics, outperforms all of the compared approaches (paired-$t$ test, $p<0.05$). The topic number is tuned on the development set, and we find that varying the number of topics does not impact performance too much. Meanwhile, both our system and \newcite{dasgupta-kumar-ravi:2013:ACL2013} produce better JSD scores than the two variants of the \newcite{Lin:2011:CSF:2002472.2002537} system, which implies the effectiveness of the dispersion function. 
We further examine the effectiveness of each component that contributes to the objective function (Section~\ref{subsec:objective}), and the results are shown in Table~\ref{tab:JSD_yahoo_main} (right).

\vspace{-2mm}
\begin{table}[ht]
    {\fontsize{8.5}{8.5}\selectfont
    \begin{minipage}{.3\linewidth}
		\begin{tabular}{|l|c|c|}
    		\hline
           &\multicolumn{2}{|c|}{\textbf{Length}}\\
	       & 100  & 200\\ \hline
		Best answer & 0.3858 & - \\ \hline
		\newcite{Lin:2011:CSF:2002472.2002537}& 0.3398 & 0.2008 \\ \hline
		\newcite{Lin:2011:CSF:2002472.2002537} + q & 0.3379 & 0.1988\\ \hline
		\newcite{dasgupta-kumar-ravi:2013:ACL2013}& 0.3316 & 0.1939\\ \hline
		Our system & \textbf{0.3017} & \textbf{0.1758}\\ \hline
		\end{tabular}
	\end{minipage}%
	\hspace{25mm}
	\begin{minipage}{.7\linewidth}
		\begin{tabular}{|l|c|c|}
    		\hline
		& \textbf{JSD$_{100}$} & \textbf{JSD$_{200}$}\\ \hline
		Rel(evance) & 0.3424 & 0.2053\\ \hline
		Rel + Aut(hor) & 0.3375 & 0.2040\\ \hline
		Rel + Aut + TM (Topic Models) & 0.3366 & 0.2033 \\ \hline
		Rel + Aut + TM + Pol(arity)& 0.3309 & 0.1983\\ \hline
		Rel + Aut + TM + Pol + Cont(ent Coverage)& 0.3102& 0.1851\\ \hline
		Rel + Aut + TM + Pol + Cont + Disp(ersion) & \textbf{0.3017}& \textbf{0.1758}\\ \hline
		\end{tabular}
	\end{minipage}
	}	
	
	\vspace{-3mm}
    \caption{\fontsize{11}{11}\selectfont  [{\bf Left}] Summaries evaluated by Jensen-Shannon divergence (JSD) on Yahoo Answer for summaries of 100 words and 200 words. The average length of the best answer is 102.70. [{\bf Right}] Value addition of each component in the objective function. The JSD on each line is statistically significantly lower than the JSD on the previous ($\alpha=0.05$).
    }
    \vspace{-2mm}
	\label{tab:JSD_yahoo_main}	
	
\end{table}

\noindent
\textbf{Human Evaluation.} 
Human evaluation for Yahoo! Answers is carried out on Amazon Mechanical Turk\footnote{\url{https://www.mturk.com/mturk/}} with carefully designed tasks (or ``HITs''). Turkers are presented summaries from different systems in a random order, and asked to provide two rankings, one for overall quality and the other for information diversity. We indicate that informativeness and non-redundancy are desirable for quality; however, Turkers are allowed to consider other desiderata, such as coherence or responsiveness, and write down those when they submit the answers. 
Here we believe that ranking the summaries is easier than evaluating each summary in isolation~\cite{Lerman:2009:SSE:1609067.1609124}.

We randomly select 100 questions from our test set, each of which is evaluated by 4 distinct Turkers located in United States. 40 HITs are thus created, each containing 10 different questions. 
Four system summaries (best answer, \newcite{dasgupta-kumar-ravi:2013:ACL2013}, and our system with 100 and 200 words respectively) are displayed along with one noisy summary (i.e. irrelevant to the question) per question in random order.\footnote{Note that we aim to compare results with the gold-standard best answers of about 100 words. The evaluation of the 200-word summaries is provided only as an additional data-point.} We reject Turkers' HITs if they rank the noisy summary higher than any other. Two duplicate questions are added to test intra-annotator agreement. We reject HITs if Turkers produced inconsistent rankings for both duplicate questions. 
A total of 137 submissions of which 40 HITs pass the above quality filters.

Turkers of all accepted submissions report themselves as native English speakers. An inter-rater agreement of Fleiss' $\kappa$ of $0.28$ (fair agreement~\cite{Landis_Koch_1977}) is computed for quality ranking and $\kappa$ is $0.43$ (moderate agreement) for diversity ranking. Table~\ref{tab:amt_yahoo} shows the percentage of times a particular method is picked as the best summary, and the macro-/micro-average rank of a method, for both overall quality and information diversity. Macro-average is computed by first averaging the ranks per question and then averaging across all questions.

For overall quality, our system with a 200 word limit is selected as the best in 44.6\% of the evaluations. It outperforms the best answer (31.9\%) significantly, which suggests that our system summary covers relevant information that is not contained in the best answer. 
Our system with a length constraint of 100 words is chosen as the best for quality 12.5\% times while that of \newcite{dasgupta-kumar-ravi:2013:ACL2013} is chosen 11.0\% of the time. 
Our system is also voted as the best summary for diversity in 78.7\% of the evaluations. More interestingly, both of our systems, with 100 words and 200 words, outperform the best answer and~\newcite{dasgupta-kumar-ravi:2013:ACL2013} for average ranking (both overall quality and information diversity) significantly by using Wilcoxon signed-rank test ($p<0.05$). 
When we check the reasons given by Turkers, we found that people usually prefer our summaries due to ``helpful suggestions that covered many options" or being ``balanced with different opinions". When Turks prefer the best answers, they mostly stress on coherence and responsiveness.
Sample summaries from all the systems are displayed in Figure~\ref{fig:sample_summmary_yahoo}.

\vspace{-3mm}
\begin{table*}[ht]
\centering
    {\fontsize{8.5}{8.5}\selectfont
	\begin{tabular}{|l|c|c|c|c|c|c|c|}
    \hline
	&\textbf{Length of Summary}&\multicolumn{3}{|c|}{\textbf{Overall Quality}}&\multicolumn{3}{|c|}{\textbf{Information Diversity}}\\
	& & \%  & \multicolumn{2}{|c|}{Average Rank}& \%  & \multicolumn{2}{|c|}{Average Rank}\\
    & &Best & Macro & Micro & Best& Macro & Micro \\ \hline
    Best answer & 102.70 & 31.9\%& 2.68& 2.69 & 9.6\%& 3.27& 3.29\\ \hline
    \hline
    \newcite{dasgupta-kumar-ravi:2013:ACL2013}  & \multirow{2}{*}{100} & 11.0\%& 2.84& 2.83 & 5.0\%& 2.95&2.94\\ 
    Our system  && 12.5\% & \textbf{2.50}$^{\ast}$& \textbf{2.50}$^{\ast}$ & 6.7\%& \textbf{2.43}$^{\ast}$ & \textbf{2.43}$^{\ast}$\\ \hline
    \hline
    Our system &200& \textbf{44.6\%}& \textbf{1.98}$^{\ast}$ & \textbf{1.98}$^{\ast}$ & \textbf{78.7}\% & \textbf{1.35}$^{\ast}$ & \textbf{1.34}$^{\ast}$\\ \hline
	\end{tabular}
	}
	\vspace{-3mm}
    \caption{\fontsize{11}{11}\selectfont  Human evaluation on Yahoo! Answer Data. \textbf{Boldface} implies statistically significance compared to other results in the same columns using paired-$t$ test. Both of our systems are ranked higher (i.e. numbers in \textbf{bold} with $^{\ast}$) than the best answers voted by Yahoo! users and system summaries from \newcite{dasgupta-kumar-ravi:2013:ACL2013}.
    }
	\vspace{-5mm}
	\label{tab:amt_yahoo}
\end{table*}


\begin{figure}[ht]
    {\fontsize{7.5}{8}\selectfont
    \begin{tabular}{|p{155mm}|}
    \hline
    \underline{\textbf{Question}}: What is the long term effect of piracy on the music and film industry?\\ \hline \hline
    \underline{\textbf{\newcite{dasgupta-kumar-ravi:2013:ACL2013}}} (Qty Rank=2.75 Div. Rank=2.5):\\
    $\bullet$In short, I don't think the music industry in particular will ever enjoy the huge profits of the 90's.\\
    $\bullet$Please-People in those businesses make millions of dollars as it is !! I don't think piracy hurts them at all !!!\\
    $\bullet$The other thing will be music and movies with less quality.\\
    $\bullet$It’s a big gray area, I don’t see anything wrong with burning a mix cd or a cd for a friend so long as you’re not selling them for profit.\\
    $\bullet$By removing the profitability of music \& film companies, piracy takes away their motivation to produce new music \& movies.\\ \hline \hline
 	
	\underline{\textbf{Our system (100 words)}} (Qty Rank=2.25 Div. Rank=2.25):\\
	$\bullet$Rising costs for movies and music. The other thing will be music and movies with less quality.\\
    $\bullet$Now, with piracy, there isn't the willingness to take chances.\\
    $\bullet$But it's also like the person put the effort into it and they aren't getting paid. It's a big gray area, I don't see anything wrong with burning a mix cd or a cd for a friend so long as you're not selling them for profit.\\
    $\bullet$It is forcing them to rework their business model, which is a good thing.\\ \hline \hline
        
    \underline{\textbf{Our system (200 words)}} (Qty. Rank=2.25, Div Rank=1.25):\\
	$\bullet$Rising costs for movies and music. The other thing will be music and movies with less quality.\\
	$\bullet$Now, with piracy, there isn't the willingness to take chances. American Idol is the result of this. .... The real problem here is that the mainstream music will become even tighter. Record labels will not won't to go far from what is currently like by the majority.\\
    $\bullet$I hate when people who have billions of dollars whine about not having more money. But it's also like the person put the effort into it and they aren't getting paid ... I don't see anything wrong with burning a mix cd or a cd for a friend ....\\
	$\bullet$It is forcing them to rework their business model, which is a good thing.\\
	$\bullet$By removing the profitability of music \& film companies, piracy takes away their motivation to produce new music \& movies.    
	\\ \hline
	\end{tabular}
	}
	\vspace{-4mm}
    \caption{\fontsize{11}{11}\selectfont Sample summaries from \newcite{dasgupta-kumar-ravi:2013:ACL2013}, and our systems (100 words and 200 words). Sentences from separate bullets ($\bullet$) are partial answers from different users.}
    \vspace{-3mm}
    \label{fig:sample_summmary_yahoo}

\end{figure}

\subsection{Blog Summarization}
\vspace{-1mm}
\label{subsec:blog}

\noindent
\textbf{Automatic Evaluation.}
We use the ROUGE~\cite{Lin:2003:AES:1073445.1073465} software with standard options to automatically evaluate summaries with reference to the human labeled nuggets as those are available for this task. ROUGE-2 measures bigram overlap and ROUGE-SU4 measures the overlap of unigram and skip-bigram separated by up to four words. 
We use the ranker trained on Yahoo! data to produce relevance ordering, and adopt the system parameters from Section~\ref{sec:mturk}. 
Table~\ref{tab:rouge_tac_main} (left) shows that our system outperforms the best system in TAC'08 with highest ROUGE-2 score~\cite{uiuc}, the two baselines (TFIDF+Lexicon, and our ranker), \newcite{Lin:2011:CSF:2002472.2002537}, and \newcite{dasgupta-kumar-ravi:2013:ACL2013}.


\vspace{-3mm}
\begin{table}[ht]
    {\fontsize{8.5}{8.5}\selectfont
 	\begin{minipage}{.5\linewidth}
		\begin{tabular}{|l|c|c|c|}
    		\hline
		&\textbf{ROUGE-2} & \textbf{ROUGE-SU4} & \textbf{JSD} \\ \hline
		Best system in TAC'08 & 0.2923 & 0.3766 & 0.3286\\ \hline
		TFIDF + Lexicon&0.3069 & 0.3876 & 0.2429\\ \hline
		Ranker (ListNet) & 0.3200 & 0.3960 & 0.2293\\ \hline
		\newcite{Lin:2011:CSF:2002472.2002537}&0.2732 & 0.3582 & 0.2330\\ \hline
		\newcite{Lin:2011:CSF:2002472.2002537} + q &0.2852 &0.3700 & 0.2349\\ \hline
		\newcite{dasgupta-kumar-ravi:2013:ACL2013} &0.2618 & 0.3500 & 0.2370\\ \hline
		Our system & \textbf{0.3234} & \textbf{0.3978} & \textit{0.2258}\\ \hline
		\end{tabular}
	\end{minipage}%
	\hspace{15mm}
	\begin{minipage}{.5\linewidth}
		\begin{tabular}{|c|c|}
	    \hline
		&\textbf{Pyramid F-score}\\ \hline
		Best system in TAC'08& 0.2225\\ \hline
		\newcite{Lin:2011:CSF:2002472.2002537}& 0.2790\\ \hline
		Our system & \textbf{0.3620}\\ \hline
		\end{tabular}
	\end{minipage}
	}
	\vspace{-4mm}
    \caption{\fontsize{11}{11}\selectfont Results on TAC'08 dataset. [\textbf{Left}] Our system has significant better ROUGE scores than all the other systems except our ranker (paired-$t$ test, $p<0.05$). We also achieve the best JS divergence. [\textbf{Right}] Human evaluation with Pyramid F-score. Our system significantly outperforms the others.} 
    \vspace{-3mm}
	\label{tab:rouge_tac_main}
\end{table}

%


\noindent
\textbf{Human Evaluation.}
For human evaluation, we use the standard Pyramid F-score used in the TAC'08 opinion summarization track with $\beta=3$~\cite{Dang:2008b}. 
In the TAC task, systems are allowed to return up to 7,000 non-white characters for each question. Since the TAC metric favors recall we do not produce summaries shorter than 7,000 characters. 
%
We ask two human judges to evaluate our system along with the one that got the highest Pyramid F-score in the TAC'08 and \newcite{Lin:2011:CSF:2002472.2002537}. Cohen's $\kappa$ for inter-annotator agreement is $0.68$ (substantial). While we did not explicitly evaluate non-redundancy, both of our judges report that our system summaries contain less redundant information.

\vspace{-1mm}
\subsection{Further Discussion}

\vspace{-3mm}
\begin{table}[ht]
    {\fontsize{8.5}{8.5}\selectfont
    \begin{minipage}{.5\linewidth}
		\begin{tabular}{|l|c|c|c|c|}
    		\hline
    		\multicolumn{5}{|c|}{\textbf{Yahoo! Answer}}\\ \hline
		&\multicolumn{2}{|c|}{\textsc{Dispersion}$_{sum}$}&\multicolumn{2}{|c|}{\textsc{Dispersion}$_{min}$}\\ \hline
		\textsc{Dissimi} & Cont$_{tfidf}$ & Cont$_{sem}$ & Cont$_{tfidf}$ & Cont$_{sem}$\\ \hline
		\textit{Semantic} & 0.3143& 0.324 3& 0.3129 & 0.3232\\ \hline
		\textit{Topical} & 0.3101 & 0.3202 & 0.3106 & 0.3209\\ \hline
		\textit{Lexical} & \textbf{0.3017} & 0.3147 & 0.3071 & 0.3172\\ \hline
		\end{tabular}
    \end{minipage}%
    \hspace{2mm}
	\begin{minipage}{.5\linewidth}
		\begin{tabular}{|l|c|c|c|c|}
	    \hline
    		\multicolumn{5}{|c|}{\textbf{TAC 2008}}\\ \hline
		&\multicolumn{2}{|c|}{\textsc{Dispersion}$_{sum}$}&\multicolumn{2}{|c|}{\textsc{Dispersion}$_{min}$}\\ \hline
	\textsc{Dissimi} & Cont$_{tfidf}$ & Cont$_{sem}$ & Cont$_{tfidf}$ & Cont$_{sem}$\\ \hline
		\textit{Semantic} & 0.2216& 0.2169& 0.2772 & 0.2579\\ \hline
		\textit{Topical} & 0.2128& 0.2090& \textbf{0.3234} & 0.3056\\ \hline		
		\textit{Lexical} & 0.2167& 0.2129& 0.3117& 0.3160\\ \hline
		\end{tabular}
	\end{minipage}   
    }
	\vspace{-3mm}
    \caption{\fontsize{11}{11}\selectfont Effect of different dispersion functions, content coverage, and dissimilarity metrics on our system. 
[{\bf Left}] JSD values for different combinations on Yahoo! data, using LDA with 100 topics. All systems are significantly different from each other at significance level $\alpha = 0.05$. Systems using summation of distances for dispersion function ($h_{sum}$) uniformly outperform the ones using minimum distance ($h_{min}$). [{\bf Right}] ROUGE scores of different choices for TAC 2008 data. All systems use LDA with 40 topics. The parameters of our systems are adopted from the ones tuned on Yahoo! Answers.}
    \vspace{-3mm}
	\label{tab:JSD_diff_combination}	
\end{table}
Given that the text similarity metrics and dispersion functions play important roles in the framework, we further study the effectiveness of different content coverage functions (Cosine using TFIDF vs.\ Semantic), dispersion functions ($h_{sum}$ vs. $h_{min}$), and dissimilarity metrics used in dispersion functions (Semantic vs.\ Topical vs.\ Lexical). Results on Yahoo! Answer (Table~\ref{tab:JSD_diff_combination} (left)) show that systems using summation of distances for dispersion functions ($h_{sum}$) uniformly outperform the ones using minimum distance ($h_{min}$). Meanwhile, Cosine using TFIDF is better at measuring content coverage than WordNet-based semantic measurement, and this may due to the limited coverage of WordNet on verbs. This is also true for dissimilarity metrics. 
Results on blog data (Table~\ref{tab:JSD_diff_combination} (right)), however, show that using minimum distance for dispersion produces better results. This indicates that optimal dispersion function varies by genre. Topical-based dissimilarity also marginally outperforms the other two metrics in blog data.


\vspace{-1mm}
\section{Conclusion}
\vspace{-1mm}
We propose a submodular function-based opinion summarization framework. Tested on community QA and blog summarization, our approach outperforms state-of-the-art methods that are also based on submodularity in both automatic evaluation and human evaluation. Our framework is capable of including statistically learned sentence relevance and encouraging the summary to cover diverse topics. We also study different metrics on text similarity estimation and their effect on summarization.

\bibliographystyle{acl}

\end{document}